\documentclass[10pt,twocolumn,letterpaper]{article}

\usepackage[pagenumbers]{cvpr} 

\usepackage{graphicx}
\usepackage{amsmath}
\usepackage{amssymb}
\usepackage{booktabs}
\usepackage{multirow}
\usepackage[accsupp]{axessibility}  

\usepackage[pagebackref,breaklinks,colorlinks]{hyperref}

\usepackage[capitalize]{cleveref}
\crefname{section}{Sec.}{Secs.}
\Crefname{section}{Section}{Sections}
\Crefname{table}{Table}{Tables}
\crefname{table}{Tab.}{Tabs.}

\def\cvprPaperID{10812} 
\def\confName{CVPR}
\def\confYear{2023}

\begin{document}

\title{Cross-Modal Implicit Relation Reasoning and Aligning for \\ Text-to-Image Person Retrieval}

\author{
Ding Jiang$^{1}$, \hspace{2pt} 
Mang Ye$^{1,2}$\thanks{Corresponding Author: Mang Ye (\href{mailto:yemang@whu.edu.cn}{yemang@whu.edu.cn})} \hspace{2pt}
\\ $^{1}$National Engineering Research Center for Multimedia Software, Institute of Artificial Intelligence,
\\ Hubei Key Laboratory of Multimedia and Network Communication Engineering, 
\\School of Computer Science, Wuhan University, Wuhan, China\\
$^{2}$ Hubei Luojia Laboratory, Wuhan, China\\
{\tt\small \href{https://github.com/anosorae/IRRA}{https://github.com/anosorae/IRRA}}
}

\maketitle

\begin{abstract}
  Text-to-image person retrieval aims to identify the target person based on a given textual description query.
  The primary challenge is to learn the mapping of visual and textual modalities into a common latent space. 
  Prior works have attempted to address this challenge by leveraging separately pre-trained unimodal models to extract visual and textual features. However, these approaches lack the necessary underlying alignment capabilities required to match multimodal data effectively. Besides, these works use prior information to explore explicit part alignments, which may lead to the distortion of intra-modality information. 
  To alleviate these issues, we present IRRA: a cross-modal Implicit Relation Reasoning and Aligning framework that learns relations between local visual-textual tokens and enhances global image-text matching without requiring additional prior supervision. 
  Specifically, we first design an Implicit Relation Reasoning module in a masked language modeling paradigm. This achieves cross-modal interaction by integrating the visual cues into the textual tokens with a cross-modal multimodal interaction encoder.
  Secondly, to globally align the visual and textual embeddings, Similarity Distribution Matching is proposed to minimize the KL divergence between image-text similarity distributions and the normalized label matching distributions. The proposed method achieves new state-of-the-art results on all three public datasets, with a notable margin of about 3\%-9\% for Rank-1 accuracy compared to prior methods.
\end{abstract}

\section{Introduction}
\label{sec:intro}

\begin{figure}
  \centering
  \begin{subfigure}[t]{\linewidth}
    \centerline{\includegraphics[scale=0.88]{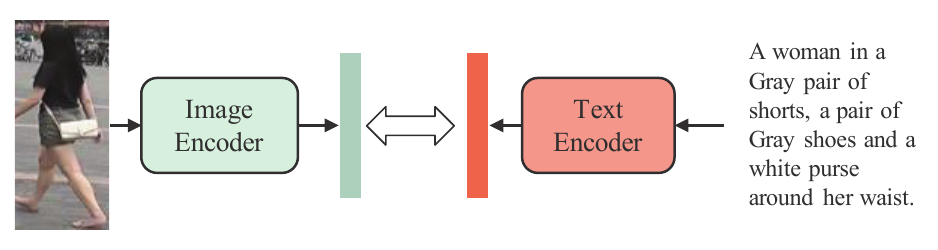}}
    \caption{Early global matching paradigm}
    \label{fig1:a}
  \end{subfigure}

  \begin{subfigure}[t]{\linewidth}
    \centerline{\includegraphics[scale=0.88]{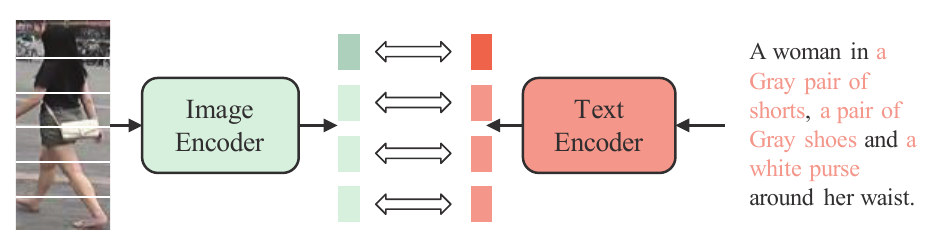}}
    \caption{Existing explicit local matching paradigm}
    \label{fig1:b}
  \end{subfigure}
  
  \begin{subfigure}[t]{\linewidth}
    \centerline{\includegraphics[scale=0.88]{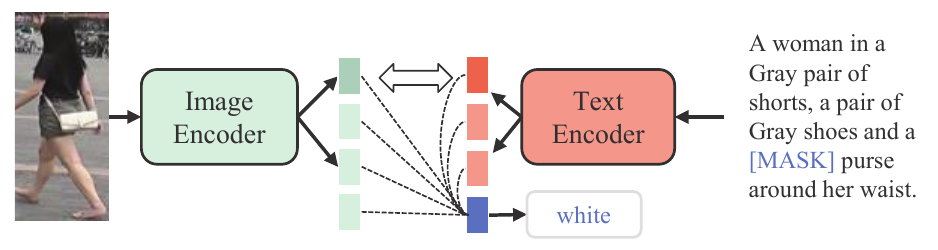}}
    \caption{Our implicit relation reasoning aided matching paradigm}
    \label{fig1:c}
  \end{subfigure}

  \begin{subfigure}[t]{\linewidth}
    \centerline{\includegraphics[width=0.95\columnwidth]{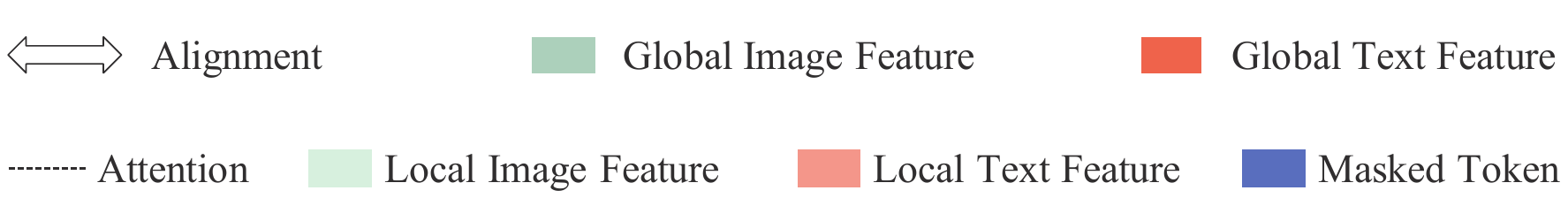}}
  \end{subfigure}
  
  \caption{Evolution of text-to-image person retrieval paradigms. (a) Early global-matching method directly align global image and text embeddings. (b) Recent local-matching method, explicitly extract and align local image and text embeddings. (c) Our implicit relation reasoning method, implicitly reasoning the relation among all local tokens to better align global image and text embeddings.}
  \label{fig1}
\end{figure}

Text-to-image person retrieval aims to retrieve a person-of-interest from a large image gallery that best matches the text description query\cite{li2017person}, which is a sub-task of both image-text retrieval\cite{lei2022loopitr, sun2021lightningdot, miech2021thinking} and image-based person re-identification (Re-ID)\cite{he2021transreid, luo2019bag, wang2022nformer}. 
Textual descriptions provide a natural and relatively comprehensive way to describe a person's attributes, and are more easily accessible than images. Text-to-image person retrieval thus received increasing attention in recent years, benefiting a variety of applications from personal photo album search to public security.

However, text-to-image person retrieval remains a challenging task due to significant intra-identity variations and modality heterogeneity between vision and language. The former challenge stems from the fact that visual appearances of an identity differ based on pose, viewpoint, illumination, and other factors, while textual description varies by arbitrary descriptive order and textual ambiguity. The latter challenge is the primary issue in cross-modal tasks and is caused by inherent representation discrepancies between vision and language.
To tackle above two challenges, the core research problem in text-to-image person retrieval is to explore better ways to extract discriminative feature representations and to design better cross-modal matching methods to align images and texts into a joint embedding space. 
Early \textit{global-matching} methods\cite{zhang2018deep, zheng2020dual} aligned images and texts into a joint embedding space by designing cross-modal matching loss functions (\cref{fig1} (a)). Typically, these approaches learned cross-modal alignments by using matching losses only at the end of the network, failing to achieve sufficient modality interaction in middle-level layers, which are crucial to bridge the feature-level modality gap.
Therefore, some later methods\cite{jing2020pose, wang2020vitaa, chen2022tipcb, ding2021semantically} introduced the practice of \textit{local-matching} by building the correspondence between the body parts and the textual entities (\cref{fig1} (b)). 
Although this local matching strategy benefits retrieval performance, it introduces unavoidable noise and uncertainty in the retrieval process. 
Besides, the strategy requires extracting and storing multiple local part representations of images and texts, computing pairwise similarity between all those representations during inference. These resource-demanding properties limit their applicability for practical large-scale scenarios. 

In this paper, we present IRRA: a \textit{cross-modal Implicit Relation Reasoning and Aligning} framework, which performs global alignment with the aid of cross-modal implicit local relation learning. 
Unlike previous methods that heavily rely on explicit fine-grained local alignment, our approach implicitly utilizes fine-grained information to enhance global alignment without requiring any additional supervision and inference costs (\cref{fig1} (c)).
Specifically, we design an Implicit Relation Reasoning module that effectively builds relations between visual and textual representations through self- and cross-attention mechanisms.
This fused representation is then utilized to perform masked language modeling (MLM) task to achieve effective implicit inter-modal and intra-modal fine-grained relation learning.
MLM is generally utilized during the pre-training stage of vision-language pre-training (VLP)\cite{lu2019vilbert,chen2020uniter,su2019vl,li2021align,dou2022empirical}. In this work, we make the first attempt to demonstrate the effectiveness of MLM in downstream fine-tuning tasks. Our main innovation is the design of a multimodal interaction encoder that can efficiently fuse visual and textual representations, align cross-modal fine-grained features through the MLM task. This design helps the backbone network to extract more discriminative global image-text representations without requiring additional supervision.
 
To guide the image-text matching, commonly used loss functions include ranking loss and cross-modal projection matching (CMPM) \cite{zhang2018deep} loss. Compared to ranking loss, the CMPM loss does not require the selection of specific triplets or margin parameter tuning. It exhibits great stability with varying batch sizes, making it widely  used in text-to-image person retrieval\cite{yan2022clip, shu2022see,chen2022tipcb}. However, we found that the projection in CMPM can be regarded as a variable weight that adjusts the distribution of softmax output logits, similar to the temperature parameter\cite{hinton2015distilling} for knowledge distillation. Nevertheless, limited by the varying projection length, CMPM therefore cannot precisely control the projection probability distribution, making it difficult to focus on hard-negative samples during model updates.
To explore more effective cross-modal matching objective, we further propose an image-text similarity distribution matching (SDM) loss.
The SDM loss minimizes the KL divergence between the normalized image-text similarity score distributions and the normalized ground truth label matching distributions. 
Additionally, we introduce a temperature hyperparameter to precisely control the similarity distribution compactness, which enables the model updates focus on hard-negative samples and effectively enlarges the variance between non-matching pairs and the correlation between matching pairs.

To address the limitations of separate pre-trained models on unimodal datasets, we leverage the Contrastive Language-Image Pre-training (CLIP)\cite{radford2021learning} as the initialization of our model. CLIP is pre-trained with abundant image-text pairs and has powerful underlying cross-modal alignment capabilities.
Some previous approaches\cite{han2021text,yan2022clip} have either frozen some part of parameters or introduced only CLIP's image encoder, which resulted in their inability to fully exploit CLIP's powerful capabilities in image-text matching. With the proposed IRRA, we successfully transfer the powerful knowledge directly from the pre-trained full CLIP model and continue to learn fine-grained cross-modal implicit local relations on text-to-image person retrieval datasets. In addition, compared to many recent methods\cite{chen2022tipcb,yan2022clip,shao2022learning}, IRRA is more efficient as it computes only one global image-text pair similarity score in the inference stage.
The main contributions can be summarized as follows:
\begin{itemize}
  
  \item We propose IRRA to implicitly utilize fine-grained interaction to enhance the global alignment without requiring any additional supervision and inference cost.
  
  \item We introduce a new cross-modal matching loss named image-text similarity distribution matching (SDM) loss. It directly minimizes the KL divergence between image-text similarity distributions and the normalized label matching distributions.
  
  \item We demonstrate that the full CLIP model can be applied to text-to-image person retrieval and can outperform existing state-of-the-art methods with straightforward fine-tuning. Moreover, our proposed IRR module enables fine-grained image-text relation learning, allowing IRRA to learn more discriminative image-text representations.
  
  \item Extensive experiments on three public benchmark datasets, \ie, CUHK-PEDES\cite{li2017person}, ICFG-PEDES\cite{ding2021semantically} and RSTPReid\cite{zhu2021dssl} show that IRRA consistently outperforms the state-of-the-arts by a large margin.

\end{itemize}

\section{Related work}
\textbf{Text-to-image Person Retrieval} was first introduced by Li \etal \cite{li2017person}, who proposed the first benchmark dataset, CUHK-PEDES\cite{li2017person}. 
The main challenge is how to efficiently align image and text features into a joint embedding space for fast retrieval.
Early works\cite{li2017identity,li2017person, chen2018improving} utilized VGG\cite{simonyan2014very} and LSTM\cite{hochreiter1997long} to learn representations for visual-textual modalities and then aligned them using a matching loss. Later works\cite{zhang2018deep, chen2021cross, sarafianos2019adversarial} improved the feature extraction backbone with ResNet50/101\cite{he2016deep} and BERT\cite{kenton2019bert}, as well as designed novel cross-modal matching losses to align global image-text features in a joint embedding space. More recent works\cite{zhu2021dssl, wang2020vitaa, wu2021lapscore, chen2022tipcb, wang2022caibc} extensively employs additional local feature learning branches that explicitly exploit human segmentation, body parts, color information, and text phrases. There is also some works \cite{farooq2022axm, ding2021semantically, shao2022learning, yan2022image} that implicitly performs local feature learning through attentional mechanisms. However, while these approaches have been shown to provide better retrieval results than using only global features, they also introduce additional computational complexity during inference when computing image-text similarity. The aforementioned works all use backbones pre-trained separately with unimodal data to extract visual and textual features, and then perform cross-modal alignment without exploiting the great cross-modal alignment capabilities of recently promising vision-language pre-training models.
Han \etal \cite{han2021text} first introduced a CLIP model for text-to-image person retrieval using a momentum contrastive learning framework to transfer the knowledge learned from large-scale generic image-text pairs. Later, Yan \etal \cite{yan2022clip} proposed a CLIP-driven fine-grain information excavation framework to transfer the knowledge of CLIP. 
However, they failed in directly transferring the original aligned CLIP dual-encoder to text-to-image person retrieval.
In this work, we demonstrate that the CLIP model can be easily transferred to text-to-image person retrieval and propose the IRRA to learn more discriminative image-text embeddings.

\textbf{Vision-Language Pre-training} aims to learn the semantic correspondence between vision and language modalities by pre-training on large-scale image-text pairs. Inspired by the success of Transformer-based\cite{vaswani2017attention} language model pre-training (such as BERT)\cite{kenton2019bert} and Vision Transformer (ViT)\cite{dosovitskiy2020image}, Vision-Language Pre-training (VLP) has emerged as the prevailing paradigm in learning multimodal representations, demonstrating strong results on downstream tasks such as image captioning\cite{chen2015microsoft}, image-text retrieval\cite{kiros2014unifying} and visual question answering\cite{antol2015vqa}.
Existing work on VLP can be categorized into two types: single-stream and dual-stream, depending on their model structure. In single-stream models\cite{chen2020uniter, su2019vl, kim2021vilt}, text and visual features are concatenated and then fed into a single transformer encoder. Although this architecture is more parameter-efficient as it uses the same set of parameters for both modalities, it has a slow retrieval speed during the inference stage because it needs to predict the similarity score of all possible image-text pairs. On the other hand, dual-stream models\cite{radford2021learning, jia2021scaling, dou2022empirical} use two separate encoders to extract the text and visual features independently. These two transformer encoders do not share parameters. While achieving remarkable performance on image-text retrieval tasks, dual-stream modals lack the ability to model complex interactions between vision and language for other vision-language understanding tasks.

\begin{figure*}[ht]
  \centerline{\includegraphics[scale=0.95]{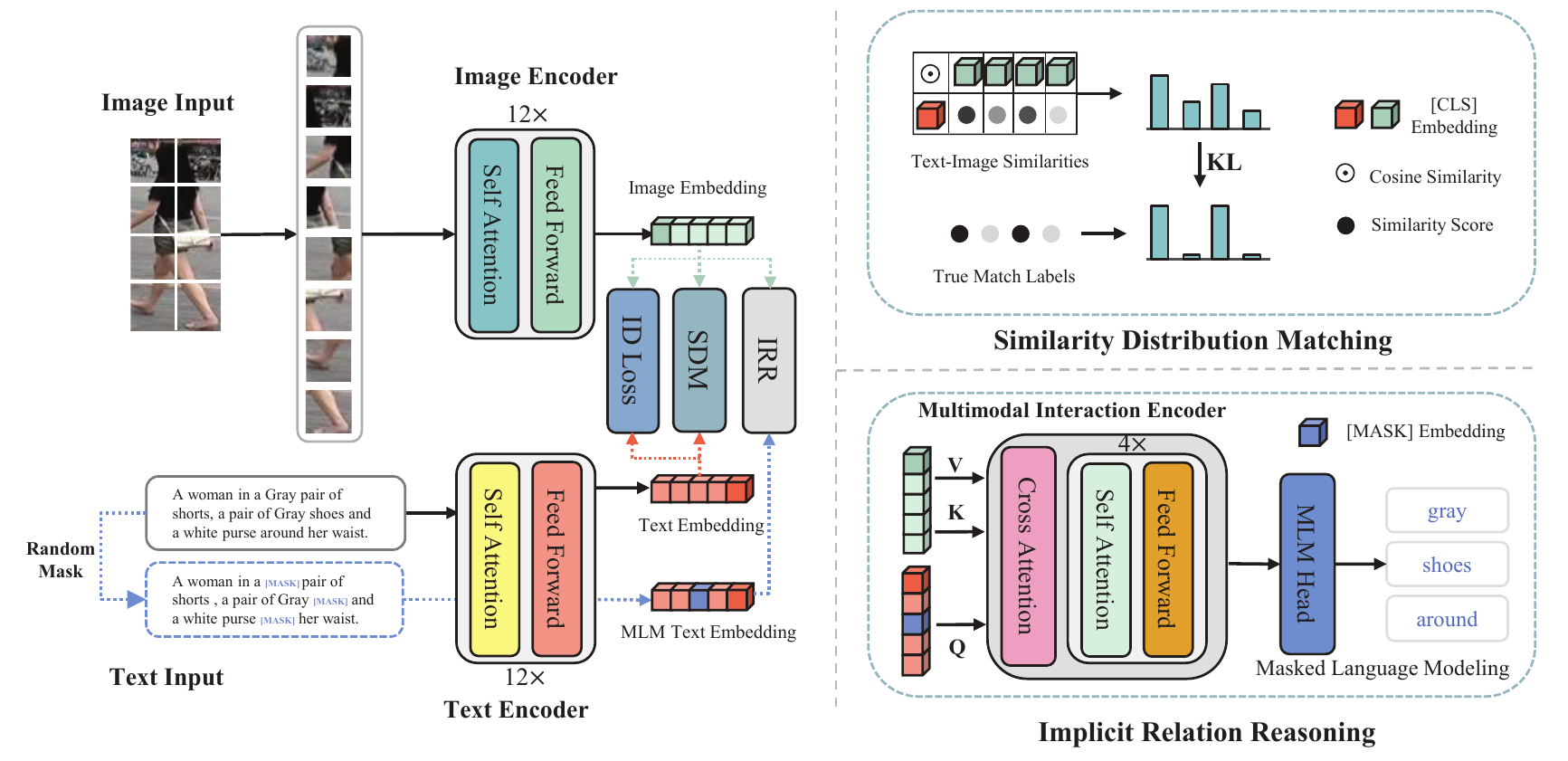}}
  \caption{\textbf{Overview of the proposed IRRA framework.} It consists of a dual-stream feature extraction backbone and three representation learning branches, \ie Implicit Relation Reasoning (IRR), Similarity Distribution Matching (SDM) and Identity Identification (ID loss). IRR aims to implicitly utilize fine-grained information to learn a discriminative global representation. SDM minimizes the KL divergence between image-text similarity score distributions and true label matching distributions, which can effectively enlarges the variance between non-matching pairs and the correlation between matching pairs. Additionally, we adopt ID loss to aggregate the feature representations of the same identity, further improving the retrieval performance. IRRA is trained end-to-end with these three tasks, and it computes only one global image-text similarity score, making it computationally efficient. Modules connected by dashed lines will be removed during inference stage.}
  \label{fig2}
\end{figure*}

\section{Method}
In this section, we present our proposed IRRA framework. The overview of IRRA is illustrated in \cref{fig2} and the details are discussed in the following subsections.

\subsection{Feature Extraction Dual-Encoder}
\label{subsec:feature}
Previous works in text-to-image person retrieval typically utilize image and text encoders that are pre-trained separately on unimodal datasets.
Inspired by the partial success of transferring knowledge from CLIP to text-image person retrieval\cite{han2021text}, we directly initialize our IRRA with the full CLIP image and text encoder to enhance its underlying cross-modal alignment capabilities.

\textbf{Image Encoder.} Given an input image $I \in R^{H\times W \times C}$, a CLIP pre-trained ViT model is adopted to obtain the image embedding. We first split $I$ into a sequence of $N = H\times W / P^2$ fixed-sized non-overlapping patches, where $P$ denotes the patch size, and then map the patch sequence to 1D tokens $ \{f_i^v\}|_{i=1}^{N} $ by a trainable linear projection. With injection of positional embedding and extra [CLS] token, the sequence of tokens $ \{f_{cls}^v, f_1^v, ..., f_N^v\} $ are input into L-layer transformer blocks to model correlations of each patch. Finally, a linear projection is adopted to map $ f_{cls}^v$ to the joint image-text embedding space, which serves as global image representation.

\textbf{Text Encoder.} For an input text $ T $, we directly use the CLIP text encoder to extract the text representation, which is a Transformer\cite{vaswani2017attention} modified by Radford \etal \cite{radford2021learning}. Following CLIP, the lower-cased byte pair encoding (BPE) with a 49152 vocab size\cite{sennrich2015neural} is firstly employed to tokenize the input text description. The text description is bracketed with [SOS] and [EOS] tokens to indicate the start and end of sequence. Then the tokenized text $ \{f_{sos}^t, f_1^t, ... f_{eos}^t\} $ are fed into the transformer and exploit correlations of each patch by masked self-attention. Finally, the highest layer of the transformer at the [EOS] token $f_{eos}^t$ is linearly projected into the image-text joint embedding space to obtain the global text representation.

\subsection{Implicit Relation Reasoning}
To fully exploit fine-grained information, it is crucial to bridge the significant modality gap between vision and language.
While most existing methods do so by explicitly aligning local features between images and text, this paper introduces a novel approach. Specifically, we use MLM to implicitly mine fine-grained relations and learn discriminative global features. 

\begin{figure}[ht]
  \centerline{\includegraphics[scale=0.9]{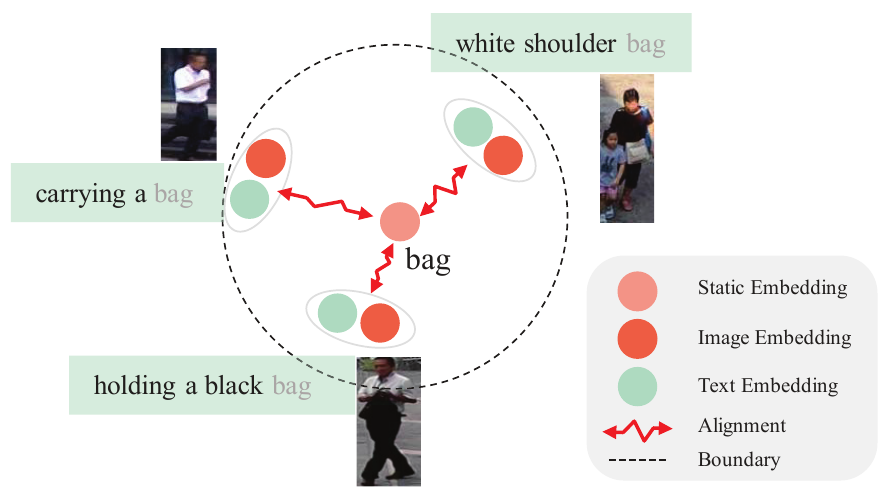}}
  \caption{Illustration of the MLM objective. MLM uses static embedding of masked textual tokens as local fine-grained keys to align image and text contextualized representations in the same context.}
  \label{fig3}
  \vspace{-3mm}
\end{figure}

\textbf{Masked Language Modeling.} Masked language modeling (MLM) was initially proposed by Taylor\cite{taylor1953cloze} in 1953, it became widely known when the BERT model adapted it as a novel pre-training task. 
In this work, We utilize MLM to predict masked textual tokens not only by the rest of unmasked textual tokens but also by the visual tokens. 
Similar to the analysis of Fu \etal \cite{fu2022contextual} in pure language pre-training, MLM optimizes two properties: (1) the alignment of image and text contextualized representations with the static embeddings of masked textual tokens, and (2) the uniformity of static embeddings in the joint embedding space. In the alignment property, sampled embeddings of masked textual tokens serve as an anchor to align images and text contextualized representations, as illustrated in \cref{fig3}. We find that such a local anchor is essential for modeling local dependencies and can implicitly utilize fine-grained local information for global feature alignment.

\begin{figure}[ht]
  \centering
  \begin{subfigure}[t]{0.38\linewidth}
    \centerline{\includegraphics[width=\linewidth]{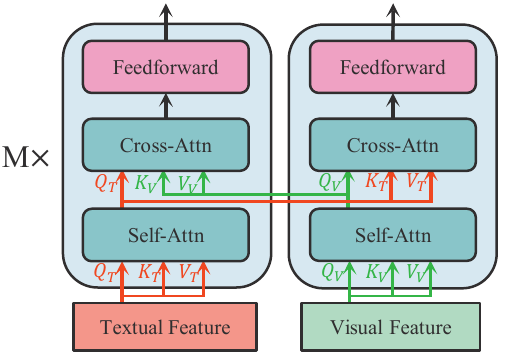}}
    \caption{Co-attention}
    \label{fig4:a}
  \end{subfigure}
  \centering
  \begin{subfigure}[t]{0.3\linewidth}
    \centerline{\includegraphics[width=\linewidth]{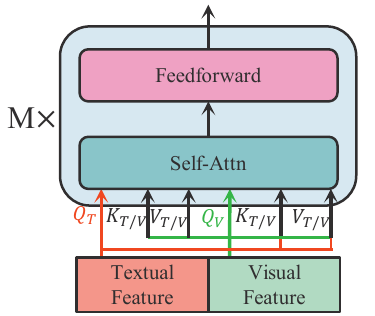}}
    \caption{Merged attention}
    \label{fig4:b}
  \end{subfigure}
  \centering
  \begin{subfigure}[t]{0.3\linewidth}
    \centerline{\includegraphics[width=\linewidth]{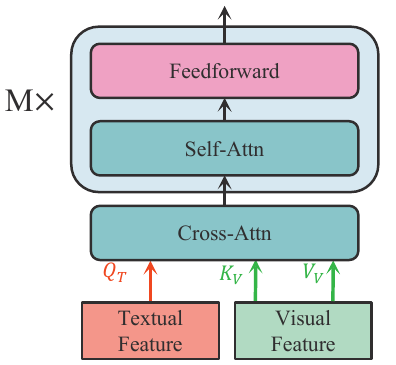}}
    \caption{Ours}
    \label{fig4:c}
  \end{subfigure}

  \caption{\textbf{Illustration of our multimodal interaction encoder and two other popular interaction modules.} (a) Co-attention, textual and visual features are fed into separate transformer blocks with self-attn and cross-attn independently to enable cross-modal interaction. (b) Merged attention, textual and visual features are concatenated together and then fed into a single transformer block. (c) Our multimodal interaction encoder, textual and visual features are first fused by a cross-attn layer and then fed into a single transformer block.}
  \label{fig4}
\end{figure}

\textbf{Multimodal Interaction Encoder.}
To achieve full interaction between image and text modalities, We design an efficient multimodal interaction encoder to fuse the image and text embeddings, compared to two other popular multimodal interaction modules \cite{hendricks2021decoupling,dou2022empirical}, our design is more computationally efficient, as illustrated in \cref{fig4}. The multimodal interaction encoder consists of a multi-head cross attention (MCA) layer and 4-layer transformer blocks.
Given an input text description $T$, we randomly mask out the text tokens with a probability of 15\% and replace them with the special token [MASK]. Following BERT, the replacements are 10\% random tokens, 10\% unchanged, and 80\% [MASK]. The masked text is defined as $\hat T$, and fed into the Text Transformer as described in \cref{subsec:feature}. Then the last hidden states $ \{h^{\hat t}_i\}_{i=1}^{L} $ and $ \{h^v_i\}_{i=1}^{N} $ of the text transformer and the vision transformer are fed into the multimodal interaction encoder jointly. In order to fuse image and masked text representations more effectively, the masked text representation $ \{h^{\hat t}_i\}_{i=1}^{L} $ served as query($\mathcal{Q}$), and the image representation $ \{h^v_i\}_{i=1}^{N} $ are served as key($\mathcal{K}$) and value($\mathcal{V} $). The full interaction between image and masked text representations can be achieved by:
\begin{equation}
  \{h^m_i\}_{i=1}^{L} = Tansformer(MCA(LN \mathcal{(Q, K, V)})),
\end{equation}
where $ \{h^m_i\}_{i=1}^{L} $ is the fused image and masked text contextualized representations, $L$ is the length of input textual tokens, $LN(\cdot)$ denotes Layer Normalization, the $MCA(\cdot)$ is the multi-head cross attention and can be realized by:
\begin{equation}
  MCA(\mathcal{Q, K, V}) = softmax(\frac{\mathcal{QK}^\top}{\sqrt{d}})\mathcal{V},
\end{equation}
where $d$ is the embedding dimension of masked tokens. 

For each masked position $\{h^m_i:i \in \mathcal{M}\}_{i=1}^{L}$, we use a multi-layer perception (MLP) classifier to predict the probability of the corresponding original tokens $ \{m_j^i\}_{j=1}^{\mathcal{|V|}}=MLP(h^m_i) $. The IRR objective can be formulated as:
\begin{equation}
  \mathcal{L}_{irr} =-\frac{1}{|\mathcal{M}| |\mathcal{V}|} \sum_{i \in \mathcal{M}} \sum_{j \in |\mathcal{V}|} y_j^i
  \log \frac{\exp (m_j^i) }{\sum_{k=1}^{|\mathcal{V}|} \exp (m_k^i)},
\end{equation}
where $\mathcal{M}$ denotes the set of masked text tokens and $\mathcal{|V|}$ is the size of vocabulary $\mathcal{V}$. $m^i$ is predicted token probability distribution and $y^i$ is a one-hot vocabulary distribution where the ground-truth token has a probability of 1.

\begin{table*}[ht]
  \centering  
  \resizebox{0.9\textwidth}{!}{
  \begin{tabular}{l|clcc|ccccc}
  \toprule
  Method                                &Type &Ref     &Image Enc.&Text Enc.     &Rank-1            &Rank-5         &Rank-10        &mAP         &mINP                \\
  \hline
  \midrule
  CMPM/C\cite{zhang2018deep}            &L    &ECCV18  &RN50      &LSTM          &49.37             &-              &79.27          &-              &-              \\
  TIMAM\cite{sarafianos2019adversarial} &G    &ICCV19  &RN101     &BERT          &54.51             &77.56          &79.27          &-              &-              \\
  ViTAA\cite{wang2020vitaa}             &L    &ECCV20  &RN50      &LSTM          &54.92             &75.18          &82.90          &51.60          &-              \\
  NAFS\cite{gao2021contextual}          &L    &arXiv21 &RN50      &BERT          &59.36             &79.13          &86.00          &54.07          &-              \\
  DSSL\cite{zhu2021dssl}                &L    &MM21    &RN50      &BERT          &59.98             &80.41          &87.56          &-              &-              \\
  SSAN\cite{ding2021semantically}       &L    &arXiv21 &RN50      &LSTM          &61.37             &80.15          &86.73          &-              &-              \\
  LapsCore\cite{wu2021lapscore}         &L    &ICCV21  &RN50      &BERT          &63.40             &-              &87.80          &-              &-              \\
  ISANet\cite{yan2022image}             &L    &arXiv22 &RN50      &LSTM          &63.92             &82.15          &87.69          &-              &-              \\
  LBUL\cite{wang2022look}               &L    &MM22    &RN50      &BERT          &64.04             &82.66          &87.22          &-              &-              \\
  Han et al.\cite{han2021text}          &G    &BMVC21  &CLIP-RN101&CLIP-Xformer  &64.08             &81.73          &88.19          &60.08          &-              \\
  SAF\cite{li2022learning}              &L    &ICASSP22&ViT-Base  &BERT          &64.13             &82.62          &88.40          &-              &-              \\
  TIPCB\cite{chen2022tipcb}             &L    &Neuro22 &RN50      &BERT          &64.26             &83.19          &89.10          &               &-              \\
  CAIBC\cite{wang2022caibc}             &L    &MM22    &RN50      &BERT          &64.43             &82.87          &88.37          &-              &-              \\
  AXM-Net\cite{farooq2022axm}           &L    &MM22    &RN50      &BERT          &64.44             &80.52          &86.77          &58.73          &-              \\
  LGUR\cite{shao2022learning}           &L    &MM22    &DeiT-Small&BERT          &65.25             &83.12          &89.00          &-              &-              \\
  IVT\cite{shu2022see}                  &G    &ECCVW22 &ViT-Base  &BERT          &65.59             &83.11          &89.21          &-              &-              \\
  CFine\cite{yan2022clip}               &L    &arXiv22 &CLIP-ViT  &BERT          &\underline{69.57} &85.93          &91.15          &-              &-              \\
  \hline
  \textbf{Baseline (CLIP-RN50)}         &G    &-       &CLIP-RN50 &CLIP-Xformer  &57.26             &78.57          &85.58          &50.88          &34.44          \\
  \textbf{Baseline (CLIP-RN101)}         &G    &-       &CLIP-RN101&CLIP-Xformer  &60.27             &80.88          &87.88          &53.93          &37.54          \\
  \textbf{Baseline (CLIP-ViT-B/16)}     &G    &-       &CLIP-ViT  &CLIP-Xformer  &68.19  &\underline{86.47} &\underline{91.47} &\underline{61.12} &\underline{44.86}\\
  \textbf{IRRA (Ours)}                 &G    &-       &CLIP-ViT  &CLIP-Xformer  &\textbf{73.38} &\textbf{89.93} &\textbf{93.71} &\textbf{66.13} &\textbf{50.24}    \\
  \bottomrule
  \end{tabular}}%
  \caption{Performance comparisons with state-of-the-art methods on CUHK-PEDES dataset. Results are ordered based on the Rank-1 accuracy. ``G" and ``L" in ``Type" column stand for global-matching/local-matching method.}
  \label{tab:1}
  \vspace{-4mm}
  \end{table*}

\subsection{Similarity Distribution Matching}
We introduce a novel cross modal matching loss termed as Similarity Distribution Matching (SDM), which incorporates the cosine similarity distributions of the $N \times N$ image-text pairs embeddings into KL divergence to associate the representations across different modalities. 

Given a mini-batch of N image-text pairs, for each image global representation $f^v_i$, we construct a set of image-text representation pairs as $\{(f^v_i, f^t_j), y_{i,j}\}_{j=1}^N $, where $y_{i,j}$ is a true matching label, $y_{i,j} = 1$ means that $(f^v_i, f^t_j)$ is a matched pair from the same identity, while $y_{i,j} = 0$ indicates the unmatched pair. Let $sim(\mathbf{u, v}) = \mathbf{u^\top v}/\|\mathbf{u}\| \|\mathbf{v}\| $ denotes the dot product between $\mathcal{L}_2$ normalized $\mathbf{u}$ and $\mathbf{v}$ (\ie cosine similarity).
Then the probability of matching pairs can be simply calculated with the following softmax function: 
\begin{equation} \label{eq4}
  p_{i,j} = \frac{\exp(sim(f_i^v, f_j^t)/\tau )}{\sum_{k=1}^{N} \exp(sim(f_i^v, f_k^t)/\tau)}, 
\end{equation}
where $\tau$ is a temperature hyperparameter which controls the probability distribution peaks. The matching probability $p_{i,j}$ can be viewed as the proportion of the cosine similarity score between $f_i^v$ and $f_j^t$ to the sum of cosine similarity score between $f_i^v$ and $\{f_j^t\}_{j=1}^{N} $ in a mini-batch. Then the SDM loss from image to text in a mini-batch is computed by :
\begin{equation} \label{eq5}
  \mathcal{L}_{i2t} = KL(\mathbf{p_i}\| \mathbf{q_i}) = \frac{1}{N} \sum_{i=1}^{N}\sum_{j=1}^{N}p_{i,j}\log(\frac{p_{i,j}}{q_{i,j} + \epsilon}), 
\end{equation}
where $\epsilon$ is a small number to avoid numerical problems, and $q_{i,j} = y_{i,j} /\sum_{k=1}^{N}y_{i,k}$ is the true matching probability. 

Symmetrically, the SDM loss from text to image $\mathcal{L}_{t2i}$ can be formulated by exchanging $f^v$ and $f^t$ in Eq.\eqref{eq4} \eqref{eq5}, and the bi-directional SDM loss is calculated by:
\begin{equation}
  \mathcal{L}_{sdm} = \mathcal{L}_{i2t} + \mathcal{L}_{t2i}. 
\end{equation}

\textbf{Optimization.}
As mentioned previously, the main objective of IRRA is to improve the learning of global image-text representations in joint embedding space. To achieve this goal, the commonly utilized ID loss\cite{zheng2020dual} is also adopted along with SDM loss and IRR loss to optimize IRRA. The ID loss is a softmax loss which classifies an image or text into distinct groups based on their identities. It explicitly considers the intra-modal distance and ensures that feature representations of the same image/text group are closely clustered together in the joint embedding space.

IRRA is trained in an end-to-end manner and the overall optimization objective for training is defined as:
\begin{equation}
  \mathcal{L} = \mathcal{L}_{irr} + \mathcal{L}_{sdm} + \mathcal{L}_{id}. 
\end{equation}

\section{Experiments}
We extensively evaluate our method on three challenging text-to-image person retrieval datasets.

\textbf{CUHK-PEDES} \cite{li2017person} is the first dataset dedicated to text-to-image person retrieval, which contains 40,206 images and 80,412 textual descriptions for 13,003 identities. Following the official data split, the training set consists of 11,003 identities, 34,054 images and 68,108 textual descriptions. The validation set and test set contain 3,078 and 3,074 images, 6158 and 6156 textual descriptions, respectively, and both of them have 1,000 identities.

\textbf{ICFG-PEDES} \cite{ding2021semantically} contains a total of 54,522 images for 4,102 identities. Each image has only one corresponding textual description. The dataset is divided into a training set and a test set, the former comprises 34,674 image-text pairs of 3,102 identities, while the latter contains 19,848 image-text pairs for the remaining 1,000 identities.

\textbf{RSTPReid} \cite{zhu2021dssl} contains 20505 images of 4,101 identities from 15 cameras. Each identity has 5 corresponding images taken by different cameras and each image is annotated with 2 textual descriptions. Following the official data split, the training, validation and test set contain 3701, 200 and 200 identities respectively.

\textbf{Evaluation Metrics.} We adopt the popular Rank-\textit{k} metrics (\textit{k}=1,5,10) as the primary evaluation metrics. Rank-\textit{k} reports the probability of finding at least one matching person image within the top-k candidate list when given a textual description as a query. In addition, for a comprehensive evaluation, we also adopt the mean Average Precision (mAP) and mean Inverse Negative Penalty(mINP)\cite{ye2021deep} as another retrieval criterion. The higher Rank-\textit{k}, mAP and mINP indicates better performance.

\textbf{Implementation Details.} IRRA consists of a pre-trained image encoder, \ie, CLIP-ViT-B/16, a pre-trained text encoder, \ie, CLIP text Transformer, and a random-initialized multimodal interaction encoder. For each layer of the multimodal interaction encoder, the hidden size and number of heads are set to 512 and 8. During training, random horizontally flipping, random crop with padding, and random erasing are employed for image data augmentation. All input images are resized to $384 \times 128$. The maximum length of the textual token sequence $L$ is set to 77. Our model is trained with Adam optimizer\cite{kingma2015adam} for 60 epochs with a learning rate initialized to $1 \times 10^{-5}$ and  cosine learning rate decay. At the beginning, we spend 5 warm-up epochs linearly increasing the learning rate from $1 \times 10^{-6}$ to $1 \times 10^{-5}$. For random-initialized modules, we set the initial learning rate to $5 \times 10^{-5}$. The temperature parameter $\tau$ in SDM loss is set to 0.02. This work is supported by Huawei MindSpore\cite{mindspore}. We perform our experiments on a single RTX3090 24GB GPU.

\subsection{Comparison with State-of-the-Art Methods}
In this section, we present comparison results with state-of-the-art methods on three public benchmark datasets.
Note that the Baseline models in \cref{tab:1} \ref{tab:2} and \ref{tab:3} denotes different CLIP models fine-tuned with the original CLIP loss (InfoNCE\cite{oord2018representation}).

\textbf{Performance Comparisons on CUHK-PEDES}
We first evaluate the proposed method on the most common benchmark, CUHK-PEDES. As shown in \cref{tab:1}, IRRA outperforms all state-of-the-art methods, achieving 73.38\% Rank-1 accuracy and 66.13\% mAP respectively. It is worth noting that our directly fine-tuned CLIP Baseline has already achieved the recent state-of-the-art method CFine\cite{yan2022clip}, with Rank-1 accuracy and mAP reaching 68.19\% and 86.47\% respectively.
In \cref{tab:1}, we annotate the feature extraction backbones ("Image Enc." and "Text Enc." column) employed by each method, and it is evident that there is a  growing demand of powerful feature extraction backbone for text-to-image person retrieval, with transformer-based backbone becoming progressively dominant. 

\textbf{Performance Comparisons on ICFG-PEDES}
The experimental results on the ICFG-PEDES dataset are reported in \cref{tab:2}. The Baseline can achieve comparable results to recent state-of-the-art methods, with 56.74\%, 75.72\% and 82.26\% on Rank-1, Rank-5 and Rank-10, respectively. Moreover, our proposed IRRA achieves 63.46\%, 80.24\% and 85.82\% on these metrics, which exceed the recent state-of-the-art local-matching method Cfine\cite{yan2022clip} by a large margin, \ie, +2.63\%, +3.69\% and +3.4\%. It is worth noting that the mINP\cite{ye2021deep} metric on ICFG-PEDES is relatively low, which indicates the inferior capability of IRRA to find the hardest matching samples.

\textbf{Performance Comparisons on RSTPReid}
We also report our experimental results on the newly released RSTPReid dataset in \cref{tab:3}. Our proposed IRRA dramatically surpass the recent global-matching method IVT\cite{shu2022see} by +13.5\%, +11.3\% and +9.4\% on Rank-1, Rank-5 and Rank-10, respectively. Compared with the recent local-matching method Cfine\cite{yan2022clip}, IRRA also achieves considerable performance gains, with the rise of +9.65\%, +8.8\% and +6.6\% on Rank-1, Rank-5 and Rank-10, respectively.

In summary, our IRRA consistently achieves the best performance for all metrics on all three benchmark datasets. This demonstrates the generalization and robustness of our proposed method.

\begin{table}[t]
  \centering  
  \resizebox{\columnwidth}{!}{
  \begin{tabular}{l|c|ccccc}
    \toprule
    Method                            &Type   &Rank-1       &Rank-5      &Rank-10       &mAP        &mINP         \\
    \hline
    Dual Path\cite{zheng2020dual}     &G      &38.99        &59.44       &68.41         &-          &-            \\
    CMPM/C\cite{zhang2018deep}        &L      &43.51        &65.44       &74.26         &-          &-            \\
    ViTAA\cite{wang2020vitaa}         &L      &50.98        &68.79       &75.78         &-          &-            \\
    SSAN\cite{ding2021semantically}   &L      &54.23        &72.63       &79.53         &-          &-            \\
    IVT\cite{shu2022see}              &G      &56.04        &73.60       &80.22         &-          &-            \\
    ISANet\cite{yan2022image}         &L      &57.73        &75.42       &81.72         &-          &-            \\
    CFine\cite{yan2022clip}           &L &\underline{60.83} &\underline{76.55} &\underline{82.42} &-  &-          \\

    \hline
    \textbf{Baseline (CLIP-RN50)}     &G      &41.46        &63.68       &73.04         &21.00      &2.46         \\
    \textbf{Baseline (CLIP-RN101)}    &G      &44.09        &66.27       &74.75         &22.59      &2.84         \\
    \textbf{Baseline (CLIP-ViT-B/16)} &G      &56.74        &75.72	     &82.26	        &31.84	    &5.03         \\
    \textbf{IRRA (Ours)}             &G &\textbf{63.46} &\textbf{80.25} &\textbf{85.82} &\textbf{38.06} & \textbf{7.93} \\
    \bottomrule
  \end{tabular}}
  \caption{Performance comparisons with state-of-the-art methods on ICFG-PEDES dataset.}
  \label{tab:2}
\end{table}

\begin{table}[ht]
  \centering  
  \resizebox{\columnwidth}{!}{
  \begin{tabular}{l|c|ccccc}
    \toprule
    Method                            &Type   &Rank-1       &Rank-5      &Rank-10       &mAP        &mINP         \\
    \hline
    DSSL\cite{zhu2021dssl}            &G      &39.05        &62.60       &73.95         &-          &-            \\
    SSAN\cite{ding2021semantically}   &L      &43.50        &67.80       &77.15         &-          &-            \\
    LBUL\cite{wang2022look}           &L      &45.55        &68.20       &77.85         &-          &-            \\
    IVT\cite{shu2022see}              &G      &46.70        &70.00       &78.80         &-          &-            \\
    CFine\cite{yan2022clip}           &L      &50.55        &72.50       &81.60         &-          &-            \\
    \hline
    \textbf{Baseline (CLIP-RN50)}     &G      &41.40        &68.55       &77.95         &31.51      &12.71        \\
    \textbf{Baseline (CLIP-RN101)}    &G      &43.45        &67.75       &78.40         &29.91      &11.18        \\
    \textbf{Baseline (CLIP-ViT-B/16)} &G &\underline{54.05}	&\underline{80.70} &\underline{88.00} &\underline{43.41} &\underline{22.31}\\
    \textbf{IRRA (Ours)}             &G &\textbf{60.20} &\textbf{81.30} &\textbf{88.20} &\textbf{47.17} & \textbf{25.28} \\
    \bottomrule
  \end{tabular}}
  \caption{Performance comparisons with state-of-the-art methods on RSTPReid dataset.}
  \label{tab:3}
  \vspace{-4mm}
\end{table}

\begin{table*}[ht]
  \centering
  \resizebox{0.9\textwidth}{!}{%
  \begin{tabular}{c|l|ccc|ccc|ccc|ccc}
  \toprule
  \multirow{2}{*}{No.} &\multirow{2}{*}{Methods} &\multicolumn{3}{c|}{Components} &\multicolumn{3}{c|}{CUHK-PEDES} &\multicolumn{3}{c|}{ICFG-PEDES} &\multicolumn{3}{c}{RSTPReid} \\ \cline{3-14} 
       &                                         &SDM &$\mathcal{L}_{id}$ &IRR      &Rank-1  &Rank-5  &Rank-10 &Rank-1  &Rank-5  &Rank-10 &Rank-1 &Rank-5 &Rank-10\\ \hline
  0    &Baseline                                 &          &          &            &68.19   &86.47   &91.47   &56.74   &75.72   &82.26   &54.05  &80.70  &88.00  \\
  1    &+$\mathcal{L}_{cmpm}$\cite{zhang2018deep}&          &          &            &59.31   &79.66   &86.11   &53.83   &72.20   &79.02   &55.40  &77.70  &85.25  \\
  2    &+SDM                                     &\checkmark&          &            &70.42	  &86.73   &92.04   &60.45   &77.88	  &83.86   &57.20  &79.90  &88.10  \\ 
  3    &+$\mathcal{L}_{id}$                      &          &\checkmark&            &65.33	  &84.05   &90.33   &53.38   &72.70   &79.70   &54.15  &76.65  &85.00  \\
  4    &+IRR                                     &          &          &\checkmark  &71.23   &88.89   &93.24   &60.96   &79.02   &84.90   &57.90 &\underline{80.85}&\underline{88.50} \\ 
  5    &+SDM+$\mathcal{L}_{id}$                  &\checkmark&\checkmark&            &70.52   &87.59   &92.12   &61.03   &78.26   &83.89   &58.65  &80.70  &87.05  \\
  6    &+SDM +IRR                                &\checkmark&          &\checkmark  &\underline{72.81}&\underline{89.31}&\underline{93.39}&\underline{63.27}&\underline{80.10}&\underline{85.77}&\underline{59.25}&79.70&88.00 \\ \hline
  7    &IRRA                                    &\checkmark&\checkmark&\checkmark  &\textbf{73.38}&\textbf{89.83}&\textbf{93.71}&\textbf{63.46}&\textbf{80.25}&\textbf{85.82}&\textbf{60.20} &\textbf{81.30} &\textbf{88.20}\\ 
  \bottomrule
  \end{tabular}%
  }
  \caption{Ablation study on each component of IRRA on CUHK-PEDES, ICFG-PEDES and RSTPReid.}
  \label{tab:4}
  \vspace{-3mm}
  \end{table*}

\subsection{Ablation Study}
In this subsection, we analyze the effectiveness of each component in the IRRA framework. Here, we adopt the CLIP-ViT-B/16 model fine-tuned with InfoNCE loss as the Baseline to facilitate the ablation study.

\textbf{Ablations on proposed components} 
To fully demonstrate the impact of different components in IRRA, we conduct a comprehensive empirical analysis on three public datasets (\ie, CUHK-PEDES\cite{li2017person}, ICFG-PEDES\cite{ding2021semantically} and RSTPReid\cite{zhu2021dssl}). The Rank-1, Rank-5, Rank-10 accuracies (\%) are reported in \cref{tab:4}.

IRR learns local relations through MLM task which can be easily integrated with other transformer-based methods to facilitate fine-grained cross-modal alignment. The efficacy of IRR is revealed via the experimental results of No.0 \vs No.4, No.2 \vs No.6 and No.5 \vs No.7. Merely adding the IRR to Baseline improves the Rank-1 accuracy by 3.04\%, 4.22\% and 3.85\% on the three datasets, respectively.
The above results clearly show that IRR module are beneficial for cross-modal matching.

To demonstrate the effectiveness of our proposed similarity distribution matching (SDM) loss, we compare it with the commonly used cross-modal projection matching (CMPM) loss \cite{zhang2018deep} (No.1 \vs No.2) on the three public datasets, the SDM loss promotes the Rank-1 accuracy of the CMPM loss by 11.11\%, 6.62\%, and 2.2\%, respectively.
Besides, replace the original InfoNCE loss with the commonly used CMPM loss (No.0 \vs No.1) does not improve the performance on text-to-image person retrieval, yet it leads to performance degradation. In contrast, the SDM loss promotes the Rank-1 accuracy of the Baseline by 2.23\%, 3.71\%, and 3.15\% on three datasets, respectively. These results demonstrate that the proposed SDM loss well aligns the features representations between the two modalities.
In addition, the experimental results of No.2 \vs No.5 and No.6 \vs No.7 demonstrate the effectiveness of the ID loss. 

\begin{table}[]
\centering
\resizebox{0.95\columnwidth}{!}{%
\begin{tabular}{l|c|c|c|c|c}
\hline
Method & Param(M) & Time(ms) & Rank-1 & Rank-5 & Rank-10 \\ \hline
\textit{Co-attn} & 33.62 & 24.30 & 73.28 & 89.04 & 93.44 \\ \hline
\textit{Merged attn} & \textbf{12.61} & 19.20 & 73.21 & 89.18 & 93.70 \\ \hline
Ours & 13.66 & \textbf{6.42} & \textbf{73.38} & \textbf{89.83} & \textbf{93.71} \\ \hline
\end{tabular}%
}
\caption{Comparisons between different Multimodal Interaction Module of IRRA on CUHK-PEDES.}
\label{tab:5}
\vspace{-4mm}
\end{table}

\textbf{Analysis of the Multimodal Interaction Encoder}
To demonstrate the advantages of our proposed Multimodal Interaction Module, we compare it with two other popular multimodal interaction modules in \cref{tab:5}. The Multimodal Interaction Module in IRR is a computationally efficient operation to fuse multimodal features, building the connection between the two modalities. We extensively compare it with \textit{Co-attn} and \textit{Merged attn} under our proposed IRRA setting, and observe slight but consistent performance gain on all Rank-\textit{k} metrics. Notably, our major advantage is the computational efficiency.

\begin{figure}[ht]
  \centerline{\includegraphics[width=0.95\columnwidth]{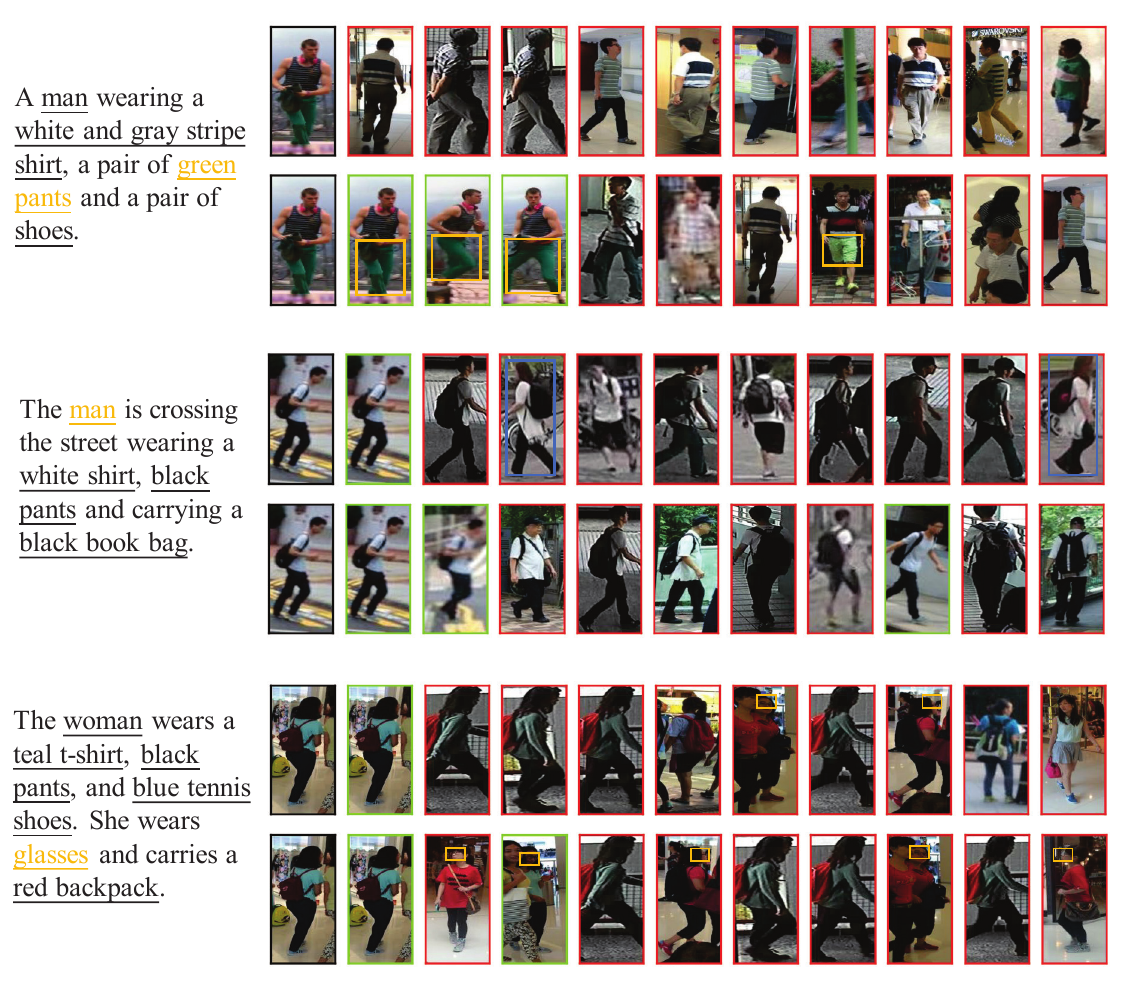}}
  \caption{Comparison of top-10 retrieved results on CUHK-PEDES between Baseline (the first row) and IRRA (the second row) for each text query. The image corresponding to query text., matched and mismatched images are marked with black, green and red rectangles, respectively.}
  \label{fig5}
  \vspace{-4mm}
\end{figure}

\subsection{Qualitative Results}
\cref{fig5} compares the top-10 retrieval results from the Baseline and our proposed IRRA. As the figure shows, IRRA achieves much more accurate retrieval results and obtains accurate retrieval results when Baseline fails to retrieve them. This is mainly due to the Implicit Relation Reasoning (IRR) modules we designed, which fully exploit fine-grained discriminative clues to distinguish different pedestrians. This is illustrated in the orange highlighted text and image regions box in \cref{fig5}. Moreover, We found that our model only learns the semantic information of the word-level but unable to understand the semantics of the phrase-level in the description text, which leads to the distortion of semantic information. This is because we only masked random single tokens in MLM, and did not perform phrase-level masking. We plan to address this issue in the future.

\section{Conclusion}
In this paper, we introduce a cross modal implicit relation reasoning and aligning framework(IRRA) to learn discriminative global image-text representations. To achieve full cross-modal interaction, we propose an Implicit Relation Reasoning module that exploits MLM to mine fine-grained relations between visual and textual tokens. 
We further propose a Similarity Distribution Matching loss to effectively enlarge the variance between non-matching pairs and the correlation between matching pairs.
These modules collaborate to align images and text into a joint embedding space. Significant performance gains on three popular benchmarks datasets prove the superiority and effectiveness of our proposed IRRA framework. We believe that the CLIP-based approach will be the future trend for text-to-image person retrieval.

\noindent\textbf{Acknowledgement.} This work is partially supported by the Key Research and Development Program of Hubei Province (2021BAA187), National Natural Science Foundation of China under Grant (62176188), Zhejiang lab (NO.2022NF0AB01), the Special Fund of Hubei Luojia Laboratory (220100015) and CAAI-Huawei MindSpore Open Fund.
{\small
\bibliographystyle{ieee_fullname}
\bibliography{egbib}
}

\end{document}